\documentclass[letterpaper]{article} 
\usepackage{aaai2026}  
\usepackage{times}  
\usepackage{helvet}  
\usepackage{booktabs} 
\usepackage{courier}  
\usepackage[hyphens]{url}  
\usepackage{graphicx} 
\usepackage{natbib}  
\usepackage{caption} 
\usepackage{algorithm} 
\usepackage{algorithmic}
\usepackage{amsfonts}
\usepackage{amsmath}

\usepackage{newfloat}
\usepackage{listings}
\DeclareCaptionStyle{ruled}{labelfont=normalfont,labelsep=colon,strut=off} 
\floatstyle{ruled}
\newfloat{listing}{tb}{lst}{}
\floatname{listing}{Listing}
\title{VSPO: Validating Semantic Pitfalls in Ontology via LLM-Based CQ Generation}
\author{
    Hyojun Choi,
    Seokju Hwang,
    Kyong-Ho Lee\thanks{Corresponding author}
}
\affiliations{
    School of Computing, Yonsei University\\
    \{gywns2184, hsjtjrwn, khlee89\}@yonsei.ac.kr
}

\begin{document}

\maketitle

\begin{abstract}

Competency Questions (CQs) play a crucial role in validating ontology design. While manually crafting CQs can be highly time-consuming and costly for ontology engineers, recent studies have explored the use of large language models (LLMs) to automate this process. However, prior approaches have largely evaluated generated CQs based on their similarity to existing datasets, which often fail to verify semantic pitfalls such as “Misusing allValuesFrom”. Since such pitfalls cannot be reliably detected through rule-based methods, we propose a novel dataset and model of Validating Semantic Pitfalls in  Ontology (VSPO) for CQ generation specifically designed to verify the semantic pitfalls. To simulate missing and misused axioms, we use LLM to generate natural language definitions of classes and properties and introduce misalignments between the definitions and the ontology by removing axioms or altering logical operators (e.g., substituting union with intersection). We then fine-tune LLaMA-3.1-8B-Instruct to generate CQs that validate these semantic discrepancies between the provided definitions and the corresponding axioms. The resulting CQs can detect a broader range of modeling errors compared to existing public datasets. Our fine-tuned model demonstrates superior performance over baselines, showing 26\% higher precision and 28.2\% higher recall than GPT-4.1 in generating CQs for pitfall validation. This research enables automatic generation of TBox-validating CQs using LLMs, significantly reducing manual effort while improving semantic alignment between ontologies and expert knowledge. To the best of our knowledge, this is the first study to target semantic pitfall validation in CQ generation using LLMs.
\end{abstract}

\begin{links}
     \link{Code}{https://github.com/Choi-Hyojun/VSPO}
\end{links}
\section{Introduction}

Ontologies are formal representations of domain knowledge, typically encoded in machine-interpretable semantic languages such as OWL~\cite{2004owl, motik2009owl}. Automated reasoners operating over ontologies can ensure logical consistency and reveal missing information, making them powerful tools for knowledge management. A widely adopted technique for supporting ontology engineering is the use of competency questions (CQs)~\cite{gruninger1995role, keet2024roles}. These natural language questions define the scope of an ontology and specify its intended requirements. CQs also facilitate the verification of whether an ontology correctly encodes the targeted knowledge~\cite{keet2024discerning, alharbi2024review}. 

\begin{figure}[t]
\centering
\includegraphics[width=0.47\textwidth]{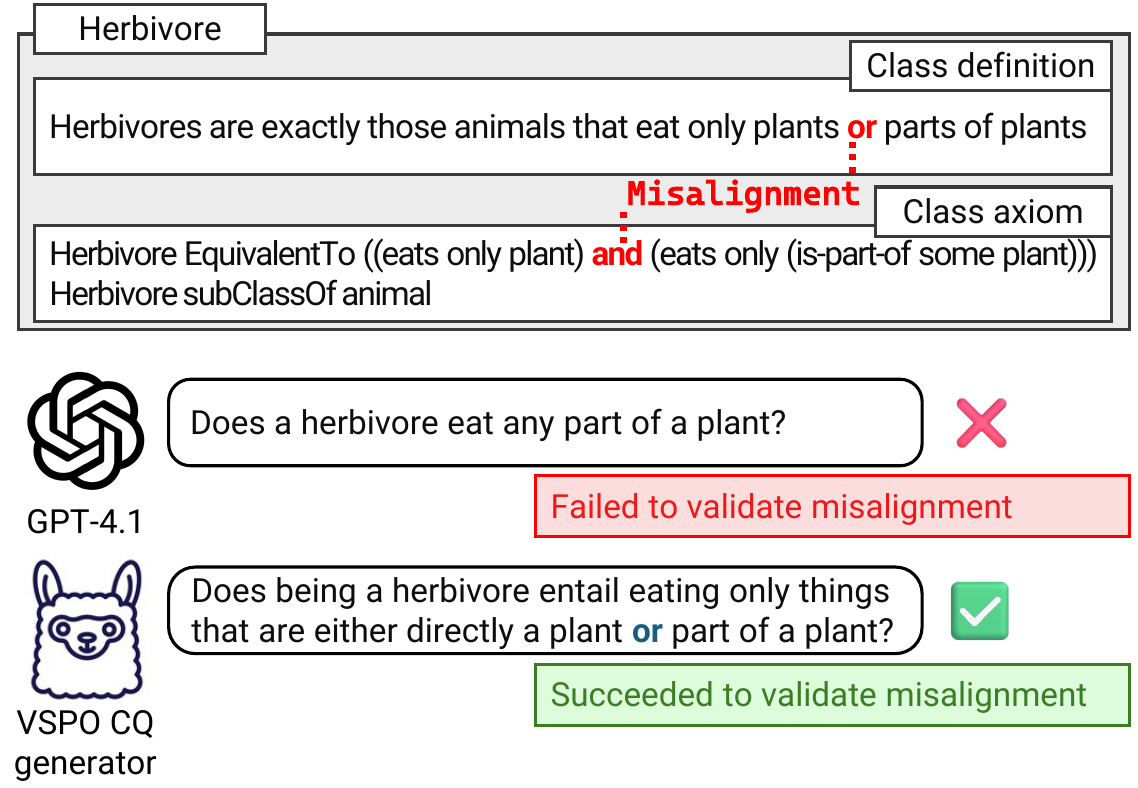}

\caption{Illustration of a semantic pitfall arising from the confusion between the class definition and axioms of Herbivore class. While our VSPO CQ generator successfully generates a CQ that validates the misalignment, GPT-4.1 fails.}
\label{fig:pitfall}
\end{figure}
The development and validation of ontologies remain labor intensive processes, as composing high-quality CQs typically requires substantial manual effort from ontology engineers, thus reflecting the overall cost associated with ontology construction.
To reduce the heavy reliance on human effort in ontology development, recent studies have started to explore the use of large language models (LLMs), such as ChatGPT, for automating various ontology engineering tasks~\cite{babaei2023llms4ol,giglou2024llms4ol,he2023exploring,mai2024llms,saeedizade2024navigating, lippolis2025bench4ke}, including the generation of CQs ~\cite{alharbi2024experiment,alharbi2024investigating,pan2024rag,rebboud2024can, alharbi2025comparative}.

Previous LLM-based approaches to CQ generation have primarily focused on reproducing human-authored CQs without explicitly addressing the underlying purpose or reasoning intent of the generated questions.
Figure \ref{fig:pitfall} illustrates an example from the African Wildlife Ontology~\cite{keet2019african} which is modified. In this case, the axiom [herbivore EquivalentTo ((eats only plant) \textbf{and} (eats only (is-part-of some plant)))] confuses union and intersection, which does not align with the definition of herbivore.
Such a pitfall creates a semantic mismatch between the natural language definition and its formal axioms, and this can lead to significant errors in ontology reasoning. However, standard ontology reasoners typically fail to detect this inconsistency.
We refer to this type of undetectable mismatch as a \textbf{semantic pitfall}, since it cannot be identified through rule-based or logical inference alone.
To address this, we generate CQs that explicitly validate these pitfalls, for instance, \textit{"Does being a herbivore entail eating only things that are either directly a plant or part of a plant?"} in Figure \ref{fig:pitfall}. In this paper, the notion of definition is used to denote any natural language definition, description, or annotation.

To develop a model capable of generating CQs that effectively verify semantic pitfalls, we construct a training dataset by introducing controlled misalignments into existing ontologies.
These misalignments are designed to simulate specific categories of semantic pitfalls, and are used to fine-tune a large language model.
We identify three primary categories of misalignment: missing axioms, undefined axioms, and misused axioms.
For example, the semantic pitfall "P10.Missing disjointness" from OOPS! (OntOlogy Pitfall Scanner!)~\cite{poveda2014oops} corresponds to the missing axiom category.
We construct a dataset that introduces controlled misalignments between ontology definitions and their corresponding axioms, and train LLM to generate CQs that explicitly address these inconsistencies. This approach enables the model to learn how to identify and validate semantic pitfalls between what is stated in a definition and what is formally encoded in the ontology.
In our experiments on misalignment CQ prediction, the fine-tuned model achieved 26\% higher precision and 28.2\% higher recall than GPT-4.1\footnote{\url{https://openai.com/index/gpt-4-1/}}.

In this work, we introduce a novel perspective on LLM-based CQ generation by shifting the focus from the similarity of the questions generated with the benchmarking CQs to their ability to discover semantic pitfalls in an ontology. Concretely, a novel dataset and model of Validating Semantic Pitfalls in an Ontology (VSPO) is proposed.
To support this, we construct a training dataset using a novel misalignment-based strategy, in which definitions and axioms are deliberately decoupled to simulate inconsistencies.
By fine-tuning a large language model on the dataset, we generate the high-quality CQs that directly target missing or misused logical structures, providing a practical step toward semi-automated ontology Tbox validation.

\section{Related Work}
In this section, we discuss previous works about ontology validation and CQ generation.
\subsection{CQs for Ontology Validation}
OOPS! models errors by empirically analyzing over 693 ontologies and compiling a live catalog of newly identified pitfalls. While some of these pitfalls can be detected using rule-based methods, other semantic pitfalls require manual human inspection. 
To generate CQs that validate such semantic pitfalls, we leverage LLMs, which are capable of substituting human-level semantic interpretation.

In Tbox validation, ~\cite{wisniewski2019analysis,potoniec2020dataset} focus on translating existing CQs into SPARQL-OWL queries for five publicly available ontologies. 
While this dataset supports useful query mapping, it primarily targets a narrow range of CQ types. For instance, existing datasets mostly focus on class-level validation. In contrast, our approach extends validation to both classes and properties, and also targets semantic pitfalls that are not addressed by existing CQs.

Since LLM-based CQ generation can easily produce a much larger number of CQs than existing datasets, we do not make a quantitative comparison with traditional TBox validation CQs.

\subsection{CQ Generation}
Since writing CQs manually is time-consuming and costly, various approaches have been explored to automate the generation of CQs. \cite{antia2023automating} propose AgOCQs, a pipeline that automatically generates CQs from domain-specific text corpora using linguistic abstraction and template matching techniques. Given a textual input corpus such as scientific articles, the system applies linguistic preprocessing and maps extracted content into abstract forms, which are then matched against CLaRO CQ templates to produce natural language CQs. The generated CQs are evaluated through expert surveys using criteria including grammaticality, answerability, relevance, and domain coverage. 

There are growing attempts to generate CQs by leveraging LLMs. \cite{rebboud2024can} prompt LLMs to generate CQs using schema inputs composed of class labels, property labels, and triples. A generated CQ is considered valid if its cosine similarity to an existing CQ exceeded a predefined threshold, and precision is reported as the proportion of valid CQs. \cite{alharbi2024experiment} use individual triples as input and incrementally elaborated the prompt to generate CQs. In a follow-up study, \cite{alharbi2024investigating} further investigate how generation behavior varies with temperature settings (0 and 0.7). They also evaluate similarity and report performance in terms of precision, recall, and F1-score. Pan et al.~\cite{pan2024rag} investigate LLMs’ ability to generate CQs using domain knowledge from scientific articles instead of any ontology information. CQs are generated using a Retrieval-Augmented Generation (RAG) approach, and precision is again computed via cosine similarity.

While the above studies evaluate similarity between generated and existing CQs, our work aims to generate CQs that validate semantic pitfalls from misalignments between natural language definitions and ontological axioms. Accordingly, our input structure incorporates both natural language class/property definitions and OWL axioms to represent potential misalignments. Moreover, unlike prior work that relies solely on inference from pre-trained models, we fine-tune an LLM and demonstrate superior performance.

\begin{figure*}[ht]
\centering
\includegraphics[width=0.9\textwidth]{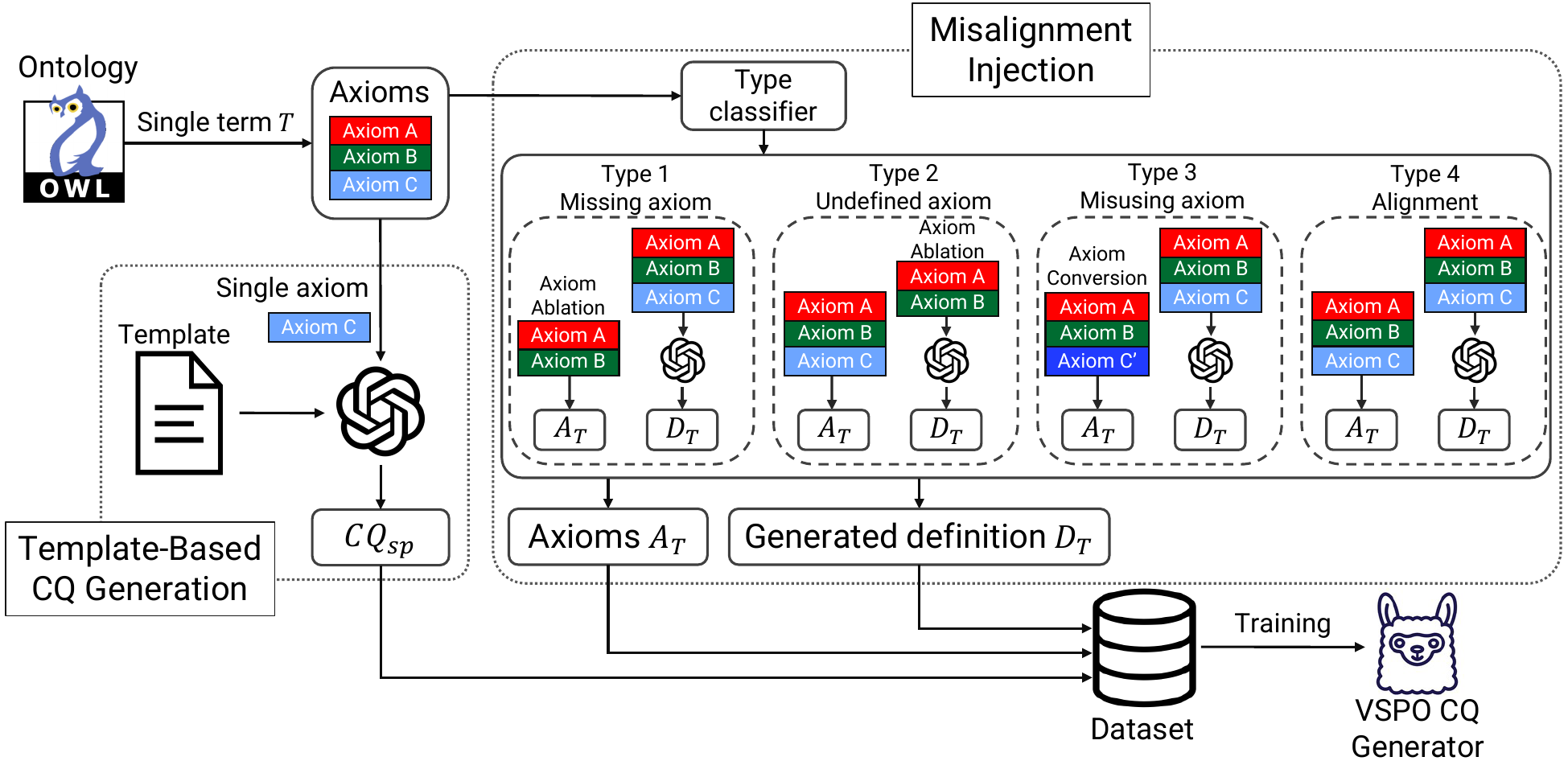}
\caption{Overall pipeline of VSPO.
An ontology term $T$ and its axioms are first classified into one of four types based on the misalignment between the axioms $A_T$ and generated definition $D_T$.
Depending on the type, a misalignment injection is applied, and template-based CQs are generated for each axiom.
These are combined into $(A_T, D_T, \mathrm{CQ}_{\text{sp}})$ triples to construct the training dataset for the VSPO CQ generator.}
\label{fig:Pipeline}
\end{figure*}

\section{Methodology}
Our objective is to generate CQs that validate, for each ontology term, the semantic pitfalls defined by OOPS!.

Let $T$ denote an \textbf{ontology term} (either a class or a property), with $A_T$ representing its axioms and $D_T$ its natural language definition. Given a pair $(A_T, D_T)$, we aim to train LLM that generates CQs ($\mathrm{CQ}_{\text{gen}}$) which maximizes semantic similarity with a reference semantic pitfall CQs ($\mathrm{CQ}_{\text{sp}}$).

Figure~\ref{fig:Pipeline} illustrates the overall pipeline.
In the dataset construction stage, we first generate CQs using axiom-specific templates per axiom. We then introduce semantic \textbf{misalignment} between the natural language definition ($D_T$) and the ontology term’s axioms ($A_T$). For each misalignment, we designate one of the previously generated CQs that validates the inconsistency as $\mathrm{CQ}_{\text{sp}}$. The resulting training dataset consists of triples in the form of $(A_T, D_T, \mathrm{CQ}_{\text{sp}})$, which are subsequently used to fine-tune the language model.

\subsection{Dataset Construction}
As a first step, we extract each individual ontology term $T$, along with all axioms in which $T$ is the subject, except for those axioms whose subject is a parent or child concept of $T$. From the extracted axioms, we construct the input data by performing a misalignment injection step, which produces both the axiom set $A_T$ and the definition $D_T$. In parallel, we apply the template-based CQ generation process to produce the target output $\mathrm{CQ}_{\text{sp}}$. Leveraging GPT-4.1, we generate these $D_T$ and $\mathrm{CQ}_{\text{sp}}$.

\subsubsection{Template-Based CQ Generation}
\label{T_based_CQ_gen}
In this stage, we generate $\mathrm{CQ}_{\text{sp}}$ based on a single axiom (e.g., axiom C as illustrated in Figure~\ref{fig:Pipeline}), following predefined prompt templates to validate the single axiom. The prompt used for CQ generation is as follows, with the reference given in \cite{alharbi2024investigating}.

\begin{quote}
\textit{
As an ontology engineer, generate a list of competency questions based on the following axiom and one-shot example. Definition of competency questions (CQs): the questions that outline the scope of ontology and provide an idea about the knowledgethat needs to be entailed in the ontology. Avoid using narrative questions + axioms. Don’t generate unnecessary text. 
Just return \{n\} distinct CQs separatedby `$|$'. Use the one-shot and known templates only as inspiration — do not copy them directly.  Rephrase and vary the structure of each CQ while maintaining its logical intent.
Generate competency questions including axioms and current template.
Template: \{template\}
\{example\}
Axiom: \{axiom\}
Generated CQs:}

\end{quote}

Each template provides the logical structure of the CQ to be generated for a given axiom, and the template varies depending on the axiom relation. For example, the template for \texttt{inverseOf} is as follows.
\begin{quote}
    {Template examples for A owl:inverseOf B axioms:}
    
    \textit{- How are property A and property B logically related in the ontology? }
    
    \textit{- If an individual C is connected to D through property A, does that imply that D is also connected to C through property B? }
    
    \textit{- What property can be inverse property of A?}
\end{quote}
For each axiom type, we manually designed 3 to 7 templates. To prevent monotonous and formulaic generation that may lead to overfitting during training, GPT-4.1 does not follow the given templates verbatim but instead generates CQ by referring only to their underlying logical structure. For each ontology term, we generate $n$ CQ per axiom, where $n=3$ in our experiments.


\subsubsection{Misalignment Injection}
At this stage, we construct a misalignment between $D_T$ and $A_T$. Since the existing annotations provide very limited information, we generate definitions based on the axioms (e.g., axiom A, B, C as illustrated in Figure~\ref{fig:Pipeline}). In the type classifier, each ontology term is randomly assigned to one of the types for which it is eligible, based on the associated axioms. Only terms with two or more associated axioms can be assigned to Type 1 or Type 2, and only terms containing axioms with ``someValuesFrom"/``allValuesFrom" or ``intersection"/``union" constructs can be assigned to Type 3. All remaining terms that do not satisfy any of these conditions are assigned only to Type 4. A detailed description of each type is provided below.

\begin{itemize}
    \item \textbf{Type 1. Missing axiom}: An axiom from $A_T$ is randomly removed, while the definition $D_T$ is generated based on the complete axiom set. For example, in Figure~\ref{fig:Pipeline}, axiom C is excluded from $A_T$ (left), while the full set of axioms is used to generate $D_T$ (right).
    
    \item \textbf{Type 2. Undefined axiom}: The axiom set $A_T$ remains complete, but one axiom is randomly removed during the generation of the definition $D_T$. For example, in Figure~\ref{fig:Pipeline}, axiom C is excluded when generating $D_T$ (right).

    \item \textbf{Type 3. Misusing axiom}: Logical constructs such as someValuesFrom/allValuesFrom or intersection/union are automatically and randomly swapped in one axiom from $A_T$, while the definition $D_T$ is generated using the original, unaltered axiom set. For example, in Figure~\ref{fig:Pipeline}, axiom C is transformed into axiom C$'$, and $A_T$ includes this modified axiom (left).

    \item \textbf{Type 4. Alignment}: No modifications are applied; both the axiom input $A_T$ and the definition $D_T$ are based on the full, consistent axiom set. For example, in Figure~\ref{fig:Pipeline}, axioms A, B, and C are used to form $A_T$ and to generate $D_T$.

\end{itemize}

The prompt used to generate definitions is as follows:

\begin{quote}
\textit{
You are an ontology engineer.
Generate a \{type\} description including information of axiom set.
The description should be concise and informative, providing a clear understanding of the \{type\}’s purpose and characteristics.
Don’t generate unnecessary text. Just generate {type} description only.
\{type\} name: \{name\}
Axiom set: \{axiom set\}
For example, \{examples\} 
Now, generate the description.
}

\end{quote}

\{type\} indicates whether the ontology term is a class or a property.

As shown in Figure \ref{fig:Pipeline}, axiom C is either removed or modified in Types 1, 2, and 3, resulting in a semantic pitfall. In such cases, the CQs generated for the axiom that underwent the semantic pitfall such as axiom C during the Template-based CQ generation stage are referred to as $\mathrm{CQ}_{\text{sp}}$. The complete set of CQs generated for all original axioms associated with the ontology term is referred to as $\mathrm{CQ}_{\text{normal}}$. The dataset consists of ($D_T$, $A_T$, $\mathrm{CQ}_{\text{sp}}$) generated in the preceding steps.

\subsection{Model Fine-Tuning}
At this stage, the LLM is trained to learn $\mathrm{CQ}_{\text{sp}}$ using the dataset constructed in the previous steps.
For Type 4, since there are no $\mathrm{CQ}_{\text{sp}}$, we randomly sample $n$ CQs from the $\mathrm{CQ}_{\text{normal}}$ and use them for training.

Since the model must be both sufficiently large to support learning and capable of generating high-quality CQs, it is essential that it has undergone instruction tuning. Accordingly, we selected LLaMA-3.1-8B-Instruct\footnote{\url{https://huggingface.co/meta-llama/Meta-Llama-3-8B-Instruct}} from the LLaMA series. To enable more efficient training, we applied the LoRA (Low-Rank Adaptation)~\cite{hu2022lora} technique. The specific implementation details are described below.

\paragraph{Implementation Details.}
We fine-tuned LLaMA-3.1-8B-Instruct using LoRA with rank $r=8$, scaling factor $\alpha=16$, and a dropout rate of 0.05. 
The model was trained for 3 epochs with an effective batch size of 4, since further training beyond 3 epochs resulted in overfitting. We used a learning rate of $3\times10^{-4}$ and bf16 precision. Training was conducted on two NVIDIA RTX 3090 GPUs using the Huggingface \texttt{Transformers}\footnote{\url{https://huggingface.co/docs/transformers/en/index}} and \texttt{PEFT}\footnote{\url{https://huggingface.co/blog/peft}} libraries.

\section{Experimental Setup}
This section provides statistics on the dataset we constructed, along with a description of the evaluation metrics.
\subsection{Datasets}

We constructed our dataset using six ontologies.
AWO~\cite{keet2019african} is an educational ontology in which most class annotations are informal descriptions, with a few derived from Wikipedia.
SWO~\cite{malone2014software} is a biomedical software ontology whose class annotations were authored using the Populous tool, although these annotations are not publicly available.
Stuff~\cite{keet2014core} is a materials ontology based on three chemistry textbooks.
Dem@Care~\cite{dasiopoulou2012demcare} is an ontology for dementia-related patient care, consisting of five OWL files.
OntoDT~\cite{panov2016generic} was constructed based on the resource titled General Purpose Datatypes.
The Pizza ontology~\cite{rector2004owl} is an educational OWL ontology that models various types of pizzas, toppings, ingredients, and their relationships.

The number of classes and properties used from each ontology is summarized in Table~\ref{tab:Number_of_Classes/Properties_in_Total}.
Except for SWO, we used all classes and properties from each ontology.
Since SWO contains a total of 3993 classes and 56 properties, we randomly sampled 500 ontology terms to maintain balance across ontologies.

\begin{table}[h]
\centering
\resizebox{\linewidth}{!}{%
\begin{tabular}{lccccccc}
\toprule
\textbf{Ontology} & \textbf{AWO} & \textbf{Dem@Care} & \textbf{SWO} & \textbf{Stuff} & \textbf{OntoDT} & \textbf{Pizza} \\
\midrule
\# of Classes     & 27  & 255 & 490 & 61  & 402 & 99   \\
\# of Properties   & 5   & 156 & 10  & 33  & 17  & 8    \\
\midrule
\# of Total       & 32  & 411 & 500 & 94  & 419 & 107 \\
\bottomrule
\end{tabular}
}
\caption{Number of Classes/Properties in total dataset}
\label{tab:Number_of_Classes/Properties_in_Total}
\end{table}

The number of ontology terms assigned to each type category by the type classifier is shown in Table~\ref{tab:Statistics_by_Misalignment_Type}. The total dataset consists of 1,563 instances, of which 1,368 were used for training and the remaining 195 were used for evaluation as the test set.

\begin{table}[h]
\centering
\begin{tabular}{lcccc}
\toprule
 & \textbf{Type 1} & \textbf{Type 2} & \textbf{Type 3} & \textbf{Type 4} \\
\midrule
\# of Classes      & 207 & 207 & 208 & 712 \\
\# of Property   & 59  & 58  & 12  & 100 \\
\midrule
\# of Total      & 266 & 265 & 220 & 812 \\
\bottomrule
\end{tabular}
\caption{Statistics by Misalignment Type}
\label{tab:Statistics_by_Misalignment_Type}
\end{table}

\subsection{Evaluation Metrics}

Following prior studies, we evaluate the semantic similarity between CQs using cosine similarity computed by SentenceBERT~\cite{reimers2019sentence}. 
For a given ontology term, let the set of generated CQs be denoted as $\mathrm{CQ}_{\text{gen}}$ and the set of ground truth CQs as $\mathrm{CQ}_{\text{gt}}$.  
We define a generated CQ as valid if it matches any gold CQ with a cosine similarity above a given threshold $\tau$:

\begin{align}
\mathrm{Valid}(\mathrm{CQ}_{\text{sp}}) = \left\{ \mathrm{cq}_{\text{gen}} \mid \max_{\mathrm{cq}_{\text{gt}}} \cos(\mathrm{cq}_{\text{gen}}, \mathrm{cq}_{\text{gt}}) \geq \tau \right\} \\
\text{where } \mathrm{cq}_{\text{gen}} \in \mathrm{CQ}_{\text{gen}}, \mathrm{cq}_{\text{gt}} \in \mathrm{CQ}_{\text{gt}}
\end{align}

Similarly, we define the set of ground truth CQs that are successfully matched by at least one generated CQ as:

\begin{align}
{
\mathrm{Matched}(CQ_{\text{gt}}) = \left\{ \mathrm{cq}_{\text{gt}}  \mid \max_{\mathrm{cq}_{\text{gen}}} \cos\left(\mathrm{cq}_{\text{gen}}, \mathrm{cq}_{\text{gt}}\right) \geq \tau \right\}
}
\end{align}

Based on these sets, we compute precision, recall, and F1-score as follows:

\begin{align}
\text{Precision} &= \frac{|\mathrm{Valid}(\mathrm{CQ}_{\text{gen}})|}{|\mathrm{CQ}_{\text{gen}}|} \\
\text{Recall} &= \frac{|\mathrm{Matched}(\mathrm{CQ}_{\text{gt}})|}{|\mathrm{CQ}_{\text{gt}}|} \\
F1\text{-score} &= 2 \cdot \frac{\text{Precision} \cdot \text{Recall}}{\text{Precision} + \text{Recall}}
\end{align}

In our experiments, the similarity threshold $\tau$ is set to 0.7.  
The reference set $\mathrm{CQ}_{\text{gt}}$ includes both $\mathrm{CQ}_{\text{sp}}$ (misalignment-based CQs) and $\mathrm{CQ}_{\text{normal}}$ (standard axiom-based CQs).

However, evaluations based on a fixed threshold $\tau$ can be highly sensitive to the choice of that threshold.  
To address this issue, we additionally report the \textbf{maximum cosine similarity} that each generated CQ achieves against the target set $\mathrm{CQ}_{\text{gt}}$, defined for every $\mathrm{cq}_{\text{gen}} \in \mathrm{CQ}_{\text{gen}}$ as:

\begin{equation}
\text{CosSim}(\mathrm{cq}_{\text{gen}}) = \max_{\mathrm{cq}_{\text{gt}} \in \mathrm{CQ}_{\text{gt}}} \cos(\mathrm{cq}_{\text{gen}}, \mathrm{cq}_{\text{gt}})
\end{equation}

\section{Expriment Results}
For all experimental result tables, P, R, F1, and C.S. denote precision, recall, F1-score, and average maximum cosine similarity, respectively. Precision, recall, and F1-score are reported as percentages, while maximum cosine similarity is presented as a value in the range [0, 1].

\subsection{Main Results}

Table~\ref{tab:overall_performance} presents the overall performance of three models—VSPO, GPT-4.1, and LLaMA-3.1-8B-Instruct—on two $\mathrm{CQ}_{\text{gt}}$ sets: $\mathrm{CQ}_{\text{sp}}$, derived from misalignment-focused questions (Types 1–3), and $\mathrm{CQ}_{\text{normal}}$, which covers general axiom-based questions across all four types.
VSPO consistently outperforms the baselines across all metrics, achieving the highest scores for $\mathrm{CQ}_{\text{sp}}$ as well. This indicates that our model has the ability to detect semantic pitfalls that previous models fail to capture and generate appropriate CQs.

For $\mathrm{CQ}_{\text{normal}}$, VSPO again achieves the best results with an F1-score and an exceptionally high precision, demonstrating its robustness even on regular axiomatic queries.
Compared to $\mathrm{CQ}_{\text{sp}}$, the precision on $\mathrm{CQ}_{\text{normal}}$ increases substantially, while the recall decreases. This is primarily because $\mathrm{CQ}_{\text{sp}}$ is a subset of $\mathrm{CQ}_{\text{normal}}$, and the number of $\mathrm{CQ}_{\text{normal}}$ instances is significantly larger, thus resulting in lower recall.
In this context, precision serves as a more meaningful metric for evaluating performance on $\mathrm{CQ}_{\text{normal}}$.

Compared to VSPO, both GPT-4.1 and LLaMA-3.1-8B-Instruct exhibit significantly lower performance, particularly on $\mathrm{CQ}_{\text{sp}}$, indicating their limited capacity to detect subtle semantic pitfalls without explicit training on misaligned examples.
\begin{table*}
\centering
\resizebox{0.75\textwidth}{!}{%
\begin{tabular}{|c|cccc|cccc|}
\hline
\textbf{Model} & \multicolumn{4}{c|}{\textbf{$\mathrm{CQ}_{\text{sp}}$}} & \multicolumn{4}{c|}{\textbf{$\mathrm{CQ}_{\text{normal}}$}} \\
\cline{2-9}
& P & R & F1 & C.S. & P & R & F1 & C.S. \\
\hline
GPT-4.1 & 49.0 & 34.0 & 40.1 & 0.6588 & 82.1 & 27.1 & 40.7 & 0.7871 \\
LLaMA-3.1-8B-Instruct & 29.8 & 20.5 & 24.3 & 0.5927 & 58.5 & 20.2 & 30.0 & 0.7156 \\
VSPO & \textbf{75.0} & \textbf{62.2} & \textbf{68.0} & \textbf{0.7950} & \textbf{95.9} & \textbf{35.8} & \textbf{52.1} & \textbf{0.8708} \\
\hline
\end{tabular}
}
\caption{Overall Performance on $\mathrm{CQ}_{\text{sp}}$ and $\mathrm{CQ}_{\text{normal}}$. }
\label{tab:overall_performance}
\end{table*}

To further investigate performance on semantic pitfall detection, Table~\ref{tab:misalignment_types} reports model performance on $\mathrm{CQ}_{\text{sp}}$ by each misalignment type.
VSPO outperforms both GPT-4.1 and LLaMA-3.1-8B-Instruct across all three types in terms of all metrics, highlighting its robust generalization across diverse misalignment patterns.
Notably, VSPO's precision in Type 2 (83.8) and Type 3 (69.0) significantly exceeds that of the baselines, suggesting its strong ability to identify CQs that correctly target semantic pitfalls, whether arising from undefined axioms (Type 2) or misusing axioms (Type 3).
While GPT-4.1 is the state of the art among non-reasoning models, it demonstrates significant limitations in detecting semantic pitfalls and generating corresponding CQs. LLaMA-3.1-8B-Instruct showed the lowest performance across all metrics.
\begin{table*}[ht]
\centering
\resizebox{\textwidth}{!}{
\begin{tabular}{|c|cccc|cccc|cccc|}
\hline
\textbf{Model} & \multicolumn{4}{c|}{\textbf{Type 1. Missing axiom}} & \multicolumn{4}{c|}{\textbf{Type 2. Undefined axiom}} & \multicolumn{4}{c|}{\textbf{Type 3. Misusing axiom}} \\
\cline{2-13}
& P & R & F1 & C.S. & P & R & F1 & C.S. & P & R & F1 & C.S. \\
\hline
GPT-4.1 & 55.3 & 37.7 & 44.8 & 0.6835 & 45.9 & 29.7 & 36.1 & 0.6498 & 44.8 & 34.5 & 39.0 & 0.6380 \\
LLaMA-3.1-8B-Instruct & 31.6 & 19.3 & 24.0 & 0.6010 & 21.6 & 17.1 & 19.1 & 0.5875 & 37.9 & 26.4 & 31.2 & 0.5884 \\
VSPO & \textbf{71.1} & \textbf{54.4} & \textbf{61.6} & \textbf{0.7867} & \textbf{83.8} & \textbf{69.4} & \textbf{75.9} & \textbf{0.8172} & \textbf{69.0} & \textbf{63.2} & \textbf{66.0} & \textbf{0.7777} \\
\hline
\end{tabular}
}
\caption{Performance on $\mathrm{CQ}_{\text{sp}}$ by each misalignment type. }
\label{tab:misalignment_types}
\end{table*}

Interestingly, while the maximum cosine similarity of LLaMA-3.1-8B-Instruct in Table~\ref{tab:misalignment_types} remains similar across the three types, its precision, recall, and F1-score fluctuate significantly. This highlights the sensitivity of these evaluation metrics to the chosen threshold value, especially when the average cosine similarity is close to the threshold boundary. This also reflects a limitation of the evaluation metrics adopted in prior LLM-based CQ generation research.

\begin{table}
\centering
\resizebox{0.47\textwidth}{!}{%
\begin{tabular}{|c|cccc|}
\hline
\textbf{Model} & P & R & F1 & C.S. \\
\hline
GPT-4.1 & 83.5 & 54.0 & 65.6 & 0.7903 \\
LLaMA-3.1-8B-Instruct & 62.6 & 37.7 & 47.0 & 0.7316 \\
VSPO & \textbf{96.7} & \textbf{68.8} & \textbf{80.4} & \textbf{0.8722} \\
\hline
\end{tabular}
}
\caption{Performance on $\mathrm{CQ}_{\text{normal}}$ (Type 4. Alignment). }
\label{tab:type4_alignment}
\end{table}
Table~\ref{tab:type4_alignment} presents the results for $\mathrm{CQ}_{\text{normal}}$ generation in the Type 4 setting, where the given axiom and definition are fully aligned.
In this setting, VSPO achieves exceptionally high performance across all metrics, indicating that it can accurately generate CQs even when no semantic pitfall is present.
Since the baseline models are not specifically trained to target only semantic pitfalls, they tend to generate more diverse outputs. Nevertheless, our VSPO model achieved higher precision and recall.
This demonstrates that the model not only learns to detect and address inconsistencies, but also develops a strong general understanding of how to generate CQs.

\subsection{Unseen Ontology Setting Results}
\begin{table}
\centering
\resizebox{0.45\textwidth}{!}{%
\begin{tabular}{|c|cccc|}
\hline
\textbf{Unseen Ontology} & P & R & F1 & C.S. \\
\hline
AWO & 83.3 & 55.6 & 66.7 & 0.8062 \\
DEM@Care & 72.1 & 56.5 & 63.3 & 0.7660 \\
Stuff & 75.4 & 53.8 & 62.8 & 0.7900 \\
SWO & 75.3 & 62.3 & 68.2 & 0.7911 \\
OntoDT & 78.2 & 60.9 & 68.5 & 0.8118 \\
Pizza & 62.0 & 46.3 & 53.0 & 0.7610 \\
\hline
\end{tabular}
}
\caption{Ontology-wise performance on unseen ontology's $\mathrm{CQ}_{\text{sp}}$ across four evaluation metrics. Each ontology represents an unseen test case. For example, the evaluation for AWO were obtained by training on the other five ontologies and testing on AWO. This leave-one-out evaluation setting was applied to all six ontologies.}
\label{tab:Unseen_ontology}
\end{table}

To evaluate the generalizability of our approach, we assessed its performance on unseen ontologies that were excluded from the training set.

As shown in Table~\ref{tab:Unseen_ontology}, the model maintains stable performance across six diverse ontologies—AWO, DEM@Care, Stuff, SWO, OntoDT, and Pizza—demonstrating strong generalization capabilities.
Despite domain and structural variations among the ontologies, the model achieves consistently high scores across all evaluation metrics.
In particular, the cases of unseen ontologies such as AWO and OntoDT exhibit strong overall performance, while unseen Pizza ontology case shows minimal degradation.
The relatively lower performance on the Pizza ontology may be attributed to its disproportionately high number of disjointWith and subClassOf axioms compared to other ontologies, which makes it more challenging to accurately detect misalignments. 

These results suggest that our VSPO is not overfitted to specific domain knowledge and can be effectively applied to a wide range of domains.

\subsection{Case Study}

\begin{table*}[ht]
\centering
\begin{tabular}{@{}p{2cm}p{15cm}@{}}
\toprule
\textbf{Model} & \textbf{Generated CQs} \\
\midrule
GPT-4.1 &
Which animals are herbivores? \newline
Does a herbivore eat any part of a plant?\newline
Can an animal that eats only plant parts be classified as a herbivore? \\
\midrule
LLaMA-3.1-8B-Instruct &
What animals are herbivores? \newline
What do herbivores eat? \newline
Are all herbivores animals? \\
\midrule
VSPO &
Does being a herbivore entail eating only things that are either directly a plant or part of a plant? \newline
Is it correct that every herbivore eats only entities which are either plants themselves or parts of plants? \newline
Can a herbivor be defined as one who consumes only what is both a plant or a part of a plant via the ``eats" property? \\
\midrule
\textit{$\mathrm{CQ}_{\text{sp}}$} &
Is every herbivore in the ontology necessarily restricted to eating only plants or only entities that are part of a plant? \newline
Is a herbivore defined as a class whose diet consists exclusively of either plants or things that include some part of a plant? \newline
Is herbivore logically equivalent to the union of animals that eat only plants and those that eat only things with at least one part being a plant? \\
\bottomrule
\end{tabular}
\caption{Generated CQs for the herbivore class, where the axiom [Herbivore EquivalentTo ((eats only plant) and (eats only (is-part-of some plant)))] is semantically misaligned with the class definition. } 
\label{tab:case_study}
\end{table*}

Table~\ref{tab:case_study} presents a case study in which models are tasked with generating CQs to validate a misusing axiom:
[herbivore EquivalentTo ((eats only plant) \textbf{and} (eats only (is-part-of some plant)))].
The definition provided in original ontology is as follows:\textit{Herbivores are exactly those animals that eat only plants \textbf{or} parts of plants.}
By definition, plants and parts of plants should be connected by a union. However, in the ontology, they are linked by an intersection. Language models are expected to generate a CQ to validate this misalignment.


VSPO successfully generates structurally grounded questions that directly target the misusing axiom. For example, it asks, “Does being a herbivore entail eating only things that are either directly a plant \textbf{or} part of a plant?” and “Is it correct that every herbivore eats only entities which are either plants themselves \textbf{or} parts of plants?”. These questions are closely aligned with $\mathrm{CQ}_{\text{sp}}$:
“Is a herbivore defined as a class whose diet consists exclusively of either plants or things that include some part of a plant?”

In contrast, both GPT-4.1 and LLaMA-3.1-8B-Instruct attempt to generate CQs that reflect some form of ontology verification.
GPT-4.1 produces contextually grounded questions that reference the given class or axiom, while LLaMA-3.1-8B-Instruct tends to generate more generic and abstract questions.
These questions, though ontologically relevant in a broad sense and translatable into SPARQL-OWL queries, fail to explicitly capture or validate the misusing intersection and union in the property restriction.

This example highlights a critical limitation of general LLMs: while they are capable of generating syntactically valid and semantically coherent questions, they often overlook subtle logical gaps or structural omissions in the ontology.
In contrast, VSPO demonstrates a stronger capacity to detect and generate questions that verify semantic pitfalls, underscoring its effectiveness for ontology validation tasks.

\section{Conclusion}
In this study, we presented the first evaluation framework that assesses LLM-based CQ generation not by surface-level similarity to existing CQs, but by the model's ability to validate semantic pitfalls in ontologies. To enable this, we constructed a training dataset using a novel misalignment-based strategy, where definitions and axioms are deliberately decoupled to simulate semantic inconsistencies. Leveraging this dataset, we trained an LLM that learns to generate CQs specifically aimed at detecting and verifying such inconsistencies. Experimental results show that our model VSPO significantly outperforms both the LLaMA-3.1-8B-Instruct and GPT-4.1, demonstrating its effectiveness and precision in ontology validation through CQ generation.
Notably, generating highly sophisticated CQs with open-source LLMs can represent a substantial advance in ontology engineering, especially for knowledge domains that involve sensitive or security-critical information. This work contributes to overcoming the bottleneck in the development of symbolic knowledge that LLMs themselves do not possess.

Despite its promising results, this work has several limitations.
First, the number of ontologies used for dataset construction was relatively small, resulting in an imbalanced distribution of axiom types. For example, properties that express ontological characteristics using \texttt{rdf:type} were extremely rare, only 11 instances were included in total.
Second, the misalignment setting was designed by removing or altering a single axiom per ontology term. Thus, cases involving multiple simultaneous semantic pitfalls were not addressed.
Third, unlike our generated definitions, most natural language annotations in real-world ontologies are sparse or ambiguous, resulting in a gap with practical scenarios. To ensure interpretability and reliability, ontology-related generation research requires outputs grounded in precise evidence. For this reason, we used a highly refined definitions.
Lastly, although VSPO is capable of generating CQs that help validate ontological consistency, it still requires human intervention to verify the generated questions, and does not yet provide a fully end-to-end solution for automated ontology validation.

In future work, we aim to extend our framework to generate not only CQs but also their corresponding SPARQL-OWL queries.
Rather than starting from axioms or definitions alone, an alternative approach would be to first generate a query and then derive a natural language question from it.
Although such a pipeline would still require some level of human feedback or evaluation, it could substantially reduce the time and cognitive load for ontology experts by providing query-question pairs that are directly verifiable.

\bigskip

\bibliography{aaai2026}
\end{document}